\documentclass[runningheads]{llncs}
\usepackage[T1]{fontenc}
\usepackage{graphicx}
\usepackage{cite}
\usepackage{amsmath}
\usepackage{amsfonts}
\usepackage{hyperref}
\usepackage{multirow}
\usepackage{booktabs} 
\usepackage{tabularx}
\usepackage{longtable}
\usepackage{supertabular}
\usepackage{subfigure}
\usepackage{caption}
\usepackage{tcolorbox}
\usepackage[algo2e, ruled,linesnumbered]{algorithm2e}
\usepackage[export]{adjustbox}
\usepackage[normalem]{ulem}
\usepackage[misc]{ifsym}
\newcommand{\corr}{(\Letter)}
\begin{document}
\begin{sloppypar}
\title{FedHCDR: Federated Cross-Domain Recommendation with Hypergraph Signal Decoupling}
\toctitle{FedHCDR: Federated Cross-Domain Recommendation with Hypergraph Signal Decoupling}
\titlerunning{
    FedHCDR
}

%
\author{Hongyu Zhang\inst{1}\orcidID{0009-0001-0098-9184},
Dongyi Zheng\inst{2,3}\orcidID{0009-0007-7541-9999}, \\
Lin Zhong\inst{4}\orcidID{0000-0002-7410-9020}, 
Xu Yang\inst{1}\orcidID{0000-0002-5343-5302}, \\
Jiyuan Feng\inst{1,2}\orcidID{0000-0003-3052-6516},
Yunqing Feng\inst{5}, \\ and
Qing Liao\inst{1,2}\corr\orcidID{0000-0003-1012-5301}}
\tocauthor{Hongyu Zhang\inst{1}\orcidID{0009-0001-0098-9184},
Dongyi Zheng\inst{2,3}\orcidID{0009-0007-7541-9999}, \\
Lin Zhong\inst{4}\orcidID{0000-0002-7410-9020}, 
Xu Yang\inst{1}\orcidID{0000-0002-5343-5302}, \\
Jiyuan Feng\inst{1,2}\orcidID{0000-0003-3052-6516},
Yunqing Feng\inst{5}, \\ and
Qing Liao\inst{1,2}\corr\orcidID{0000-0003-1012-5301}}
\authorrunning{Zhang et al.}
%
\institute{Harbin Institute of Technology (Shenzhen), Shenzhen, China \\
\email{\{orion-orion, xuyang97, fengjy\}@stu.hit.edu.cn, liaoqing@hit.edu.cn} \and
Peng Cheng Laboratory, Shenzhen, China \and
Sun Yat-sen University, Guangzhou, China \email{zhengdy23@mail2.sysu.edu.cn} \and
Chongqing University, Chongqing, China \email{zhonglin@stu.cqu.edu.cn} \and
Shanghai Pudong Development Bank, Shanghai, China \email{fengyq5@spdb.com.cn}}
\maketitle              
%
\begin{abstract}
In recent years, Cross-Domain Recommendation (CDR) has drawn significant attention, which utilizes user data from multiple domains to enhance the recommendation performance. However, current CDR methods require sharing user data across domains, thereby violating the General Data Protection Regulation (GDPR). Consequently, numerous approaches have been proposed for Federated Cross-Domain Recommendation (FedCDR). Nevertheless, the data heterogeneity across different domains inevitably influences the overall performance of federated learning. In this study, we propose \textbf{FedHCDR}, a novel \textbf{Fed}erated \textbf{C}ross-\textbf{D}omain \textbf{R}ecommendation framework with \textbf{H}ypergraph signal decoupling. Specifically, to address the data heterogeneity across domains, we introduce an approach called hypergraph signal decoupling (HSD) to decouple the user features into domain-exclusive and domain-shared features. The approach employs high-pass and low-pass hypergraph filters to decouple domain-exclusive and domain-shared user representations, which are trained by the local-global bi-directional transfer algorithm. In addition, a hypergraph contrastive learning (HCL) module is devised to enhance the learning of domain-shared user relationship information by perturbing the user hypergraph. Extensive experiments conducted on three real-world scenarios demonstrate that FedHCDR outperforms existing baselines significantly\footnote{Code available at \href{https://github.com/orion-orion/FedHCDR}{https://github.com/orion-orion/FedHCDR}}.

\keywords{Federated learning \and Recommendation system \and Graph neural network.}
\end{abstract}
\section{Introduction}

Cross-domain recommendation (CDR) \cite{DTCDR, BiTGCF} has been widely applied to leverage information on the web by providing personalized information filtering in various real-world applications, including Amazon (an e-commerce platform) and YouTube (an online video platform). CDR can significantly enhance the performance of item recommendations for users by utilizing user rating data from various domains, under the assumption that users have similar preferences across domains. However, with the formulation of the General Data Protection Regulation (GPDR), user-item ratings are not accessible across different domains. How to provide high-quality cross-domain recommendations while satisfying privacy protection has emerged as an urgent issue.

In this paper, we focus on a problem of federated cross-domain recommendation (FedCDR) \cite{FedCDR, FPPDM++}. In this case, user-item rating interactions are considered private information that cannot be directly accessed by other domains. Although existing FedCDR methods \cite{FedCDR, PriCDR, P2FCDR, FPPDM++} can effectively solve the privacy issue in CDR, they also face the issue of data heterogeneity across domains, that is, user-item interaction data in different domains contain domain-exclusive information.
\begin{figure*}
\centering
\includegraphics[width=\linewidth, scale=0.15]{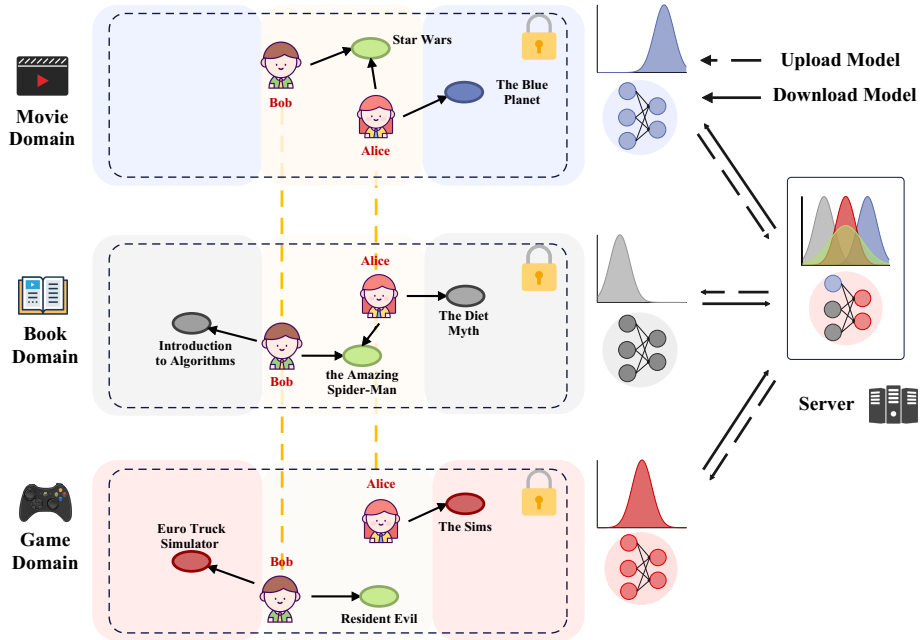}
\caption{Data heterogeneity across domains in the FedCDR scenario.}
\label{fig1}
\end{figure*}
Fig.~\ref{fig1} presents a toy example illustrating the data heterogeneity across different domains. The figure depicts that Bob and Alice interact with various types of items in each domain, including action movies and documentary movies in the Movie domain, professional books and action books in the Book domain, and action games and simulation games in the Game domain. As previously mentioned, documentary movies in the Movie domain, professional books in the Book domain, and simulation games in the Game domain can all be considered as domain-exclusive interaction information. Existing FedCDR methods \cite{FedCDR, PriCDR, P2FCDR, FPPDM++} use direct aggregation of client models or user representations to transfer knowledge, resulting in the mixing of domain-exclusive information into the global model, resulting in poor local performance of the global model (i.e., negative transfer). Therefore, it is necessary to decouple domain-shared and domain-exclusive information, and only aggregate domain-shared information to avoid negative transfer problems.

In response to the issue of data heterogeneity, we introduce a novel \textbf{Fed}erated \textbf{C}ross-\textbf{D}omain \textbf{R}ecommendation framework with \textbf{H}ypergraph signal decoupling (\textbf{FedHCDR}). This framework enables different domains to collectively train better-performing CDR models without the need to share raw user data. Specifically, inspired by the hypergraph structure~\cite{Hypergraph, DHCF} and graph spectral filtering~\cite{Graph_signal_processing, ChebNet, GCN}, we introduce a hypergraph signal decoupling method called HSD to tackle the data heterogeneity across domains. In this approach, the model of each domain is divided into a high-pass hypergraph filter and a low-pass hypergraph filter, responsible for extracting domain-exclusive and domain-shared user representations respectively. Furthermore, we devise a hypergraph contrastive learning module HCL to enhance the learning of domain-shared user relationship information by introducing perturbations to the user hypergraph. The evaluation is conducted on Amazon datasets under the federated learning setting. The experimental results demonstrate that our FedHCDR significantly enhances recommendation performance in three different FedCDR scenarios.

To summarize, our contributions are as follows:

\begin{itemize}
\item We propose a novel federated cross-domain framework FedHCDR, designed to enable different domains to train better performing CDR models collaboratively while ensuring data privacy.
\item We introduce HSD, a hypergraph signal decoupling method. HSD uses high-pass and low-pass hypergraph filters to decouple the user features into domain-exclusive and domain-shared features to address the data heterogeneity issue across domains, which are trained by the local-global bi-directional transfer algorithm.
\item We devise a hypergraph contrastive learning module HCL, which perturbs the user hypergraph to learn more effective domain-shared user relationship information.
\end{itemize}

\section{Methodology}

\subsection{Problem Formulation}
Assume there are $K$ local clients and a central server. The $k$-th client maintains its own user-item interaction data $\mathcal{D}_k = (\mathcal{U}, \mathcal{V}_k, \mathcal{E}_k)$, which forms a distinct domain, where $\mathcal{U}$ denotes the overlapped user set in all domains, $\mathcal{V}_k$ denotes the item set in domain $k$, and $\mathcal{E}_k$ denotes edge set, i.e., the set of user-item pairs in domain $k$. Additionally, there is a user-item incidence matrix $\mathbf{A}_k\in \{0, 1\}^{|\mathcal{U}|\times |\mathcal{V}_k|}$ for domain $k$, where each element $\left(\mathbf{A}_k\right)_{ij}$ describes whether user $u_i \in \mathcal{U}$ has interacted with item $v_j \in \mathcal{V}_k$ in the edge set $\mathcal{E}_k$.

\begin{figure*}
\centering
\includegraphics[height=0.62\linewidth]{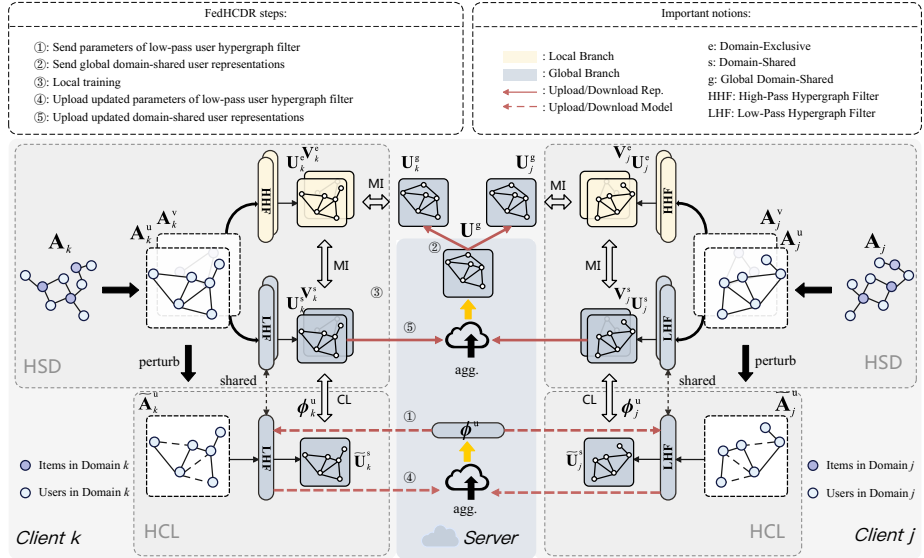}
\caption{An overview of FedHCDR.}
\label{FedHCDR-framework}
\end{figure*}

For client $k$, we first construct the user hypergraph adjacency matrix $\mathbf{A}^{\text{u}}_k$ and the item hypergraph adjacency matrix $\mathbf{A}^{\text{v}}_k$ according to the user-item incidence matrix $\mathbf{A}_k$. Subsequently, we feed the user hypergraph adjacency matrix $\mathbf{A}^{\text{u}}_k$ into the high-pass and low-pass user hypergraph filters respectively to decouple it into domain-exclusive user representations $\mathbf{U}^{\text{e}}_k$ and domain-shared user representations $\mathbf{U}^ {\text{s}}_k$, and feed the item hypergraph adjacency matrix $\mathbf{A}^{\text{v}}_k$ into high-pass and low-pass item hypergraph filters respectively to decouple it into $\mathbf{V}^{\text{e}}_k$ and $\mathbf {V}^{\text{s}}_k$. After the local model update is completed, the central server aggregates ${\{\mathbf{U}^{\mathrm{s}}_k\}}^K_{k=1}$ to obtain the global representation ${\mathbf{U}}^{\mathrm{g}}$ used in the subsequent training round. The local perturbed domain-shared user representations are denoted as $\widetilde{\mathbf{U}}_k^{\mathrm{s}}$.

Each client's local model is divided into a global branch with low-pass user/item hypergraph filters (parameterized by $\boldsymbol{\phi}^{\mathrm{u}}_k / \boldsymbol{\phi}^{\mathrm{v}}_k$), and a local branch with high-pass user/item hypergraph filters (parameterized by $\boldsymbol{\theta}_k^{\text{u}} / \boldsymbol{\theta}_k^{\text{v}}$). At the end of each training round, the server aggregates $\{\boldsymbol{\phi}_k^{\mathrm{u}}\}^{K}_{k=1}$ to derive global user low-pass hypergraph parameters $\boldsymbol{\phi}^{\mathrm{g}}$ which are then shared among clients in the subsequent training round.

\subsection{Overview of FedHCDR}

Our proposed FedHCDR, depicted in Fig.~\ref{FedHCDR-framework}, utilizes client-server federated learning architecture. Each client's model is divided into a local branch (in yellow) and a global branch (in purple). In each training round, only domain-shared user representations and model parameters are aggregated. During the test phase, both domain-exclusive and domain-shared representations are utilized together for local predictions.

\subsection{High/Low-Pass Hypergraph Filter}

\begin{figure}
    \centering
    \subfigure[Construction of the user and item hypergraph adjacency matrix.]{
    \includegraphics[height=2in]{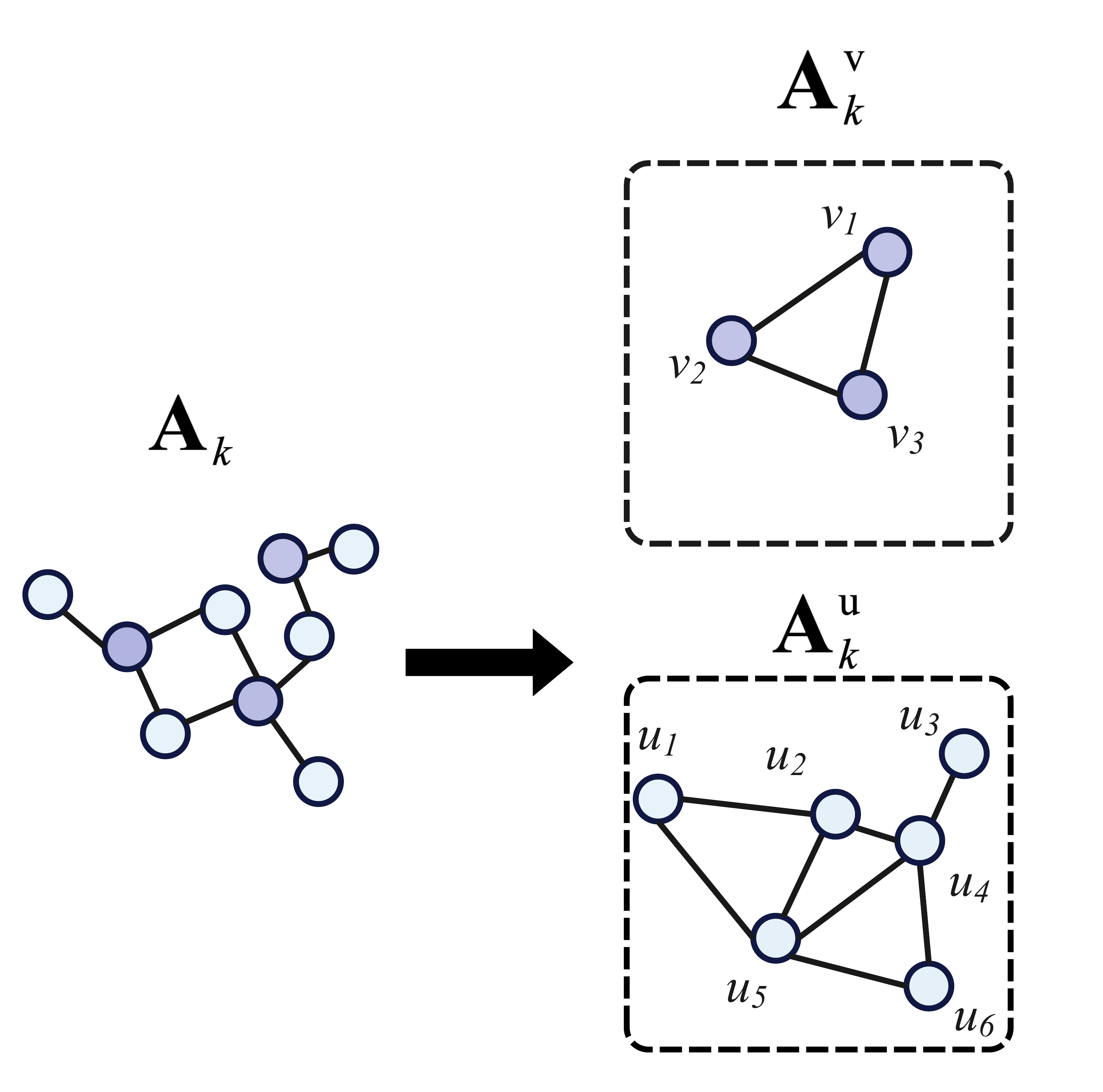}}
    \subfigure[Hypergraph random walk.]{
    \includegraphics[height=2in]{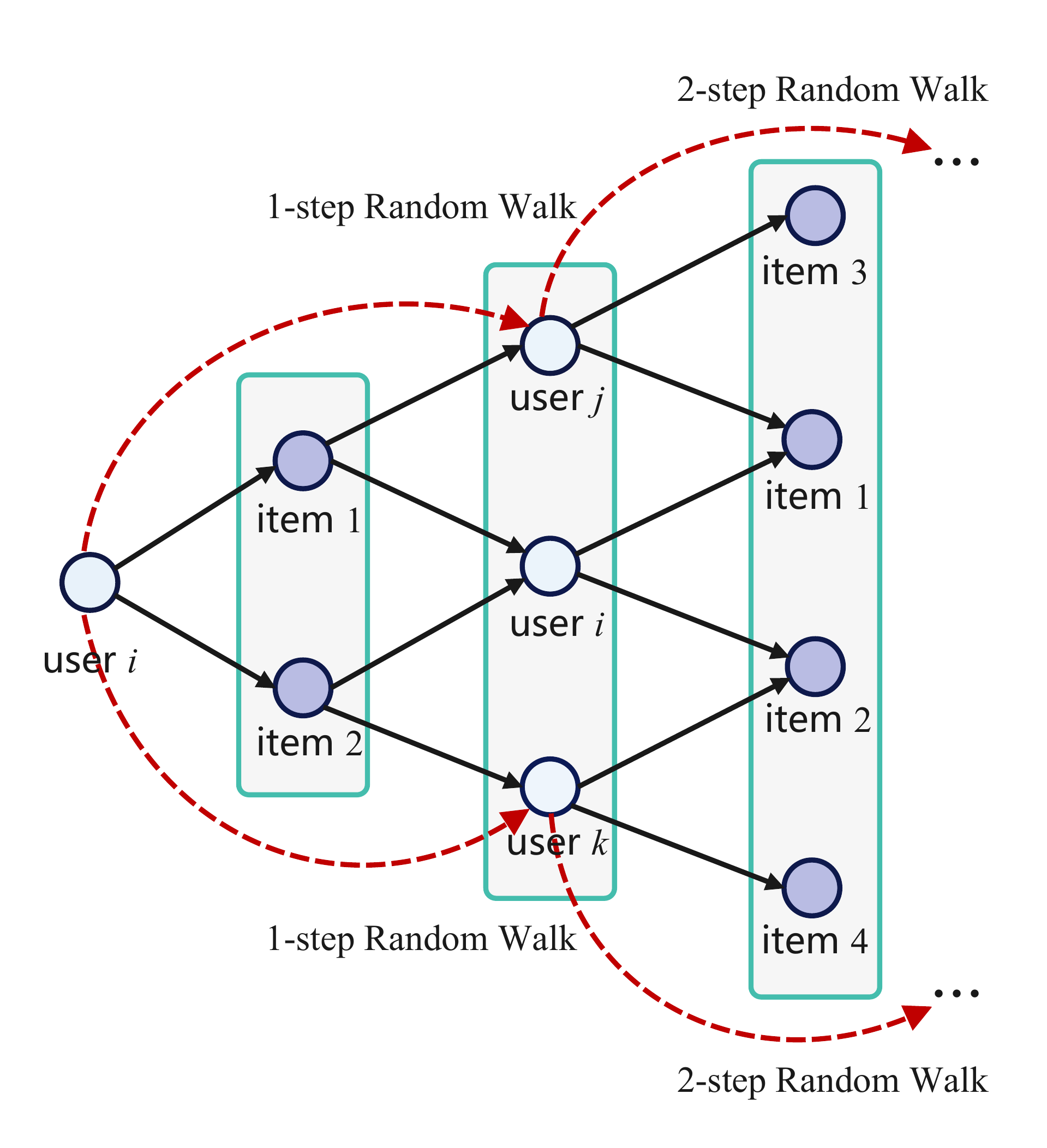}}    
    \caption{Initialization of hypergraph filter.}
    \label{HF}
\end{figure}


\subsubsection{Construction of the user and item hypergraph adjacency matrix.} First, given the user-item incidence matrix $\mathbf{A}_k$ in domain $k$, we can denote the hypergraph incidence matrices of users and items respectively as:

\begin{equation}
\mathbf{H}_k^{\mathrm{u}} =\mathbf{A}_k, \quad \mathbf{H}_k^{\mathrm{v}} =\mathbf{A}_k^T,
\end{equation}

\noindent where each element $\left(\mathbf{H}_{k}^{\text{u}}\right)_{ij}$ describes whether the vertex(user) $u_i$ belongs to the hyperedge(item) $v_j$. Then, let us denote the item popularity debiasing matrix for the user hypergraph as $\mathbf{P}_k$, and we can obtain the normalized unbiased hypergraph adjacency matrix and hypergraph Laplacian matrix as follows:

\begin{equation}
\begin{aligned}
\mathbf{A}_k^{\mathrm{u}}=\mathbf{D}_k^{\mathrm{u}, -\frac{1}{2}}\left(\mathbf{H}_k^{\mathrm{u}} \mathbf{P}_k^{\mathrm{v}} \mathbf{D}^{\mathrm{v}, -1}_k \mathbf{H}_k^{\mathrm{u}, T}-\widetilde{\mathbf{D}}_k^{\mathrm{u}}\right) \mathbf{D}_k^{\mathrm{u}, -\frac{1}{2}} \in \mathbb{R}^{|\mathcal{U}| \times|\mathcal{U}|} \\
\mathbf{L}_k^{\mathrm{u}}=\mathbf{I}-\mathbf{D}_k^{\mathrm{u}, -\frac{1}{2}}\left(\mathbf{H}_k^{\mathrm{u}} \mathbf{P}_k^{\mathrm{v}} \mathbf{D}_k^{\mathrm{v}, -1} \mathbf{H}_k^{\mathrm{u}, T}-\widetilde{\mathbf{D}}^{\mathrm{u}}_k\right) \mathbf{D}_{k}^{\mathrm{u},-\frac{1}{2}} \in \mathbb{R}^{|\mathcal{U}| \times|\mathcal{U}|},
\end{aligned}
\end{equation}
where
\begin{equation}
\begin{aligned}
& \left(\mathbf{D}_{k}^{\mathrm{u}}\right)_{uu}=\sum_{v\in \mathcal{V}_k} \left(\mathbf{H}_k^{\mathrm{u}}\right)_{uv} \left(\mathbf{P}_k^{\mathrm{v}}\right)_{vv},\quad \left(\mathbf{D}_k^{\mathrm{v}}\right)_{vv} =\sum_{u\in \mathcal{U}} \left(\mathbf{H}_k^{\mathrm{u}}\right)_{uv}
\label{joint}
\end{aligned}
\end{equation}
are the user (vertex) degree matrix and item (hyperedge) degree matrix for the user hypergraph respectively, and the item popularity debiasing matrix
\begin{equation}
\begin{aligned}
\left(\mathbf{P}_k^{\mathrm{v}}\right)_{vv}=1-\frac{\left(\mathbf{D}_k^{\mathrm{v}}\right)_{vv}}{\sum_{v\in\mathcal{V}_k} \left(\mathbf{D}_k^{\mathrm{v}}\right)_{vv}}
\label{joint}
\end{aligned}
\end{equation}
is used to remove the estimation bias caused by the long tail effect of items. Intuitively, the higher the degree (frequency of interactions) of hyperedge (item) $v$, the lower the weight $\left(\mathbf{P}_k^{\mathrm{v}}\right)_{vv}$ of it. Besides, matrix

\begin{equation}
\begin{aligned}
& \left(\widetilde{\mathbf{D}}^{\mathrm{u}}_k\right)_{uu} = \sum_{v \in \mathcal{V}_k} \left(\mathbf{H}_k^{\mathrm{u}}\right)_{uv} \frac{\left(\mathbf{P}_k^{\mathrm{v}}\right)_{vv}}{\left(\mathbf{D}_k^{\mathrm{v}}\right)_{vv}}
\label{joint}
\end{aligned}
\end{equation}
is used to ensure that the diagonal of the user hypergraph adjacency matrix $\mathbf{A}^{\text{u}}_k$ is filled with 0, that is, there is no self-loop in the user hypergraph. The construction of the user and item hypergraph adjacency matrix is shown in Fig.~\ref{HF}(a).

\subsubsection{Low-Pass hypergraph filter representation initialization.}

To make the low-pass user hypergraph filter capture smoother signals (i.e., user relationship information), we construct the Markov transition matrix $\mathbf{M}_k^u$ of the user hypergraph and perform a $\mathcal{T}$-step hypergraph random walk to obtain the initialized low-frequency user representations $\mathbf{U}_k^{\mathrm{s},(0)}$, which is shown in Fig.~\ref{HF}(b). The Markov transition matrix of the user hypergraph (the same applies to the item hypergraph) is formulated as follows:

\begin{equation}
\begin{aligned}
\mathbf{M}^{\mathrm{u}}_k =\mathbf{D}_k^{\mathrm{u}, -1}\left(\mathbf{H}_k^{\mathrm{u}} \mathbf{P}_k^{\mathrm{v}} \mathbf{D}_{k}^{\mathrm{v}, -1} \mathbf{H}_k^{\text{u}, T}-\widetilde{\mathbf{D}}_k^{\mathrm{u}}\right) \in \mathbb{R}^{|\mathcal{U}| \times|\mathcal{U}|},
\label{joint}
\end{aligned}
\end{equation}

\noindent where $\left({\mathbf{M}}^{\text{u}}_k\right)_{ij} \in [0, 1]$ indicates the probability that the random walk moves from user $i$ to user $j$ in one step.

Then for the user hypergraph, we can get the transition matrices of random walk of steps $1, 2, \cdots, \mathcal{T}$:

\begin{equation}
\begin{aligned}
\mathbf{M}^{\mathrm{u}, 1}_k, \mathbf{M}_k^{\text{u}, 2}, \ldots, \mathbf{M}_k^{\mathrm{u}, \mathcal{T}}.
\label{joint}
\end{aligned}
\end{equation}

\noindent Take diagonals of the transition matrices (where $\left(\mathbf{M}^{\text{u}, t}_k\right)_{ii}$ indicates the probability of going from  user $i$ and return to itself in exactly $t$ steps), and perform a linear transformation after concatenating, then the initialized user low-frequency representations $\mathbf{U}_k^{\mathrm{s},(0)}$ are obtained:

\begin{equation}
\begin{aligned}
&\widetilde{\mathbf{M}}_k^u=\operatorname{diag}\left(\mathbf{M}_k^{\mathrm{u}, 1}\right)\left\|\operatorname{diag}\left(\mathbf{M}_{k}^{\mathrm{u}, 2}\right), \cdots,\right\| \operatorname{diag}\left(\mathbf{M}_k^{\mathrm{u}, \mathcal{T}}\right)\\
&\mathbf{U}_k^{\mathrm{s},(0)}=f_k\left(\widetilde{\mathbf{M}}_k^u\right)
\label{joint}
\end{aligned}
\end{equation}




Note that for the remaining hypergraph filters, we directly use the EmbeddingLookup operation to obtain randomly initialized representations. For example, for the high-pass user hypergraph filter, we have:

\begin{equation}
\begin{aligned}
\mathbf{U}_k^{\mathrm{e},(0)}=\mathbf{E}_k^{\mathrm{u}}=\text{EmbeddingLookup}\left(\mathbf{X}_k^u\right),
\label{joint}
\end{aligned}
\end{equation}

\noindent where $\mathbf{X}^{\text{u}}_k\in \{0, 1\}^{|\mathcal{U}|\times |\mathcal{U}|}$ is the matrix of one-hot IDs of users. \noindent Same with $\mathbf{V}^{\text{s}, (0)}_k / \mathbf{V}^{\text{e}, (0)}_k$.

\subsubsection{Adaptive high/low-pass hypergraph filtering.} Based on the user and item hypergraph adjacency matrices $\mathbf{A}_k^u$ and $\mathbf{A}_k^v$ (The corresponding Laplacians are $\mathbf{L}_k^u$, $\mathbf{L}_k^v$), as well as the initialized user and item representations $\mathbf{U}^{\text{s}, (0)}_k / \mathbf{U}^{\text{e}, (0)}_k$, and $\mathbf{V}^{\text{s}, (0)}_k / \mathbf{V}^{\text{s}, (0)}_k$, we further perform adaptive high-pass/low-pass hypergraph filtering on user and item representations. Let us denote user and item representations after $l$ layer graph convolution as $\mathbf{U}_k^{\text{s}, (l)} / \mathbf{U}_k^{\text{e}, (l)}$ and $\mathbf{V}_k^{\text{s}, (l)} / \mathbf{V}_k^{\text{e}, (l)}$.

According to the graph signal processing theory \cite{Graph_signal_processing, ChebNet}, given a signal $\boldsymbol{x}\in\mathbb{R}^{|\mathcal{V}|}$ on a graph $\mathcal{G}=(\mathcal{V}, \mathcal{E})$ and a filter $\boldsymbol{h}$, the graph convolution/filtering operator in the spectral (frequency) domain (denoted as $*_{\mathcal{G}}$) is defined as 
\begin{equation}
\begin{aligned}
\boldsymbol{y} = \boldsymbol{x}*_{\mathcal{G}} \boldsymbol{h} = \mathbf{Q}\left((\mathbf{Q}^T \boldsymbol{x}) \odot (\mathbf{Q}^T \boldsymbol{h})\right)
\end{aligned}
\end{equation}
, where $\mathbf{Q}$ is the matrix of eigenvectors of the graph Laplacian, $\odot$ denotes the element-wise Hadamard product, and $\boldsymbol{y}$ is the output signal. In particular, we can define $\mathbf{Q}^{T}\boldsymbol{h} = \left(h(\lambda_1, ), \cdots h(\lambda_n)\right)^{T}$, where $\lambda_k$ is the $k$-th eigenvalue of graph Laplacian $\textbf{L}$ and $h(\lambda)$ is the spectral filter. Let $h(\mathbf{\Lambda}) = \text{diag} \left(h(\lambda_1), \cdots, h(\lambda_n) \right)$, then we have:

\begin{equation}
\begin{aligned}
\boldsymbol{y} =\left(\mathbf{Q} h(\mathbf{\Lambda}) \mathbf{Q}^T\right) \boldsymbol{x}= h(\mathbf{L}) \boldsymbol{x},
\end{aligned}
\end{equation}

Following the previous works \cite{ChebNet, GCN}, we adopt $K$-order Chebyshev polynomials $h(\mathbf{L}) = \sum_{k=0}^K w_k T_k(\widehat{\mathbf{L}})$ to approximate $h(\mathbf{L})$. Let $K=1, w_0=w, w_1=w, \widehat{\mathbf{L}}=\mathbf{L}_k^u - (1+\beta)\mathbf{I}$, where $\beta$ is a trainable parameter, then for the user hypergraph, the adaptive high-pass filtering operation is defined as follows:

\begin{equation}
\begin{aligned}
\boldsymbol{y} & =\left(w \mathbf{I}+w\left(\mathbf{L}_k^{\mathrm{u}}-(\beta+1) \mathbf{I}\right)\right) \boldsymbol{x} \\
& =w\left(\mathbf{L}_k^{\mathrm{u}}-\beta \mathbf{I}\right) \boldsymbol{x} \\
& =w\left((1-\beta) \mathbf{I}-\mathbf{A}_k^{\mathrm{u}}\right) \boldsymbol{x},
\end{aligned}
\end{equation}

\noindent where the hypergraph filter $h(\mathbf{L})=\mathbf{L}_k^u - \beta \mathbf{I}, h(\lambda)=\lambda - \beta$, which is a adaptive high-pass hypergraph filter. Then for the high-pass user hypergraph filter, we have the following layer-wise propagation rule:

\begin{equation}
\begin{aligned}
\mathbf{U}_k^{(\ell+1)} &=\sigma\left(\mathbf{P}^{(l)}_k \mathbf{U}_k^{(\ell)} \mathbf{W}_k^{\mathrm{u},(\ell)} + \boldsymbol{b} \right)\\
\mathbf{P}^{(l)}_k &= \left(1-\beta^{(l)}\right) \mathbf{I} - \mathbf{D}_u^{-\frac{1}{2}}\left(\mathbf{H}_k^{\mathrm{u}} \mathbf{P}_k^{\mathrm{v}} \mathbf{D}_{\mathrm{v}}^{-1} \mathbf{H}_k^{\mathrm{u}, T}-\widetilde{\mathbf{D}}_u\right) \mathbf{D}_u^{-\frac{1}{2}}
\end{aligned}
\end{equation}

Let $K=1, w_0=w, w_1=-w, \widehat{\mathbf{L}}=\mathbf{L}_k^u-(1+\beta)\mathbf{I}$, where $\beta$ is a trainable parameter, then for the user hypergraph, the adaptive low-pass filtering operation is defined as follows:

\begin{equation}
\begin{aligned}
\boldsymbol{y} & =\left(w \mathbf{I}-w\left(\mathbf{L}_k^{\mathrm{u}}-(\beta+1) \mathbf{I}\right)\right) \boldsymbol{x} \\
& =w\left((2+\beta) \mathbf{I}-\mathbf{L}_k^{\mathrm{u}}\right) \boldsymbol{x} \\
& =w\left((1+\beta) \mathbf{I}+\mathbf{A}_k^{\mathrm{u}}\right) \boldsymbol{x},
\end{aligned}
\end{equation}

\noindent where the hypergraph filter $h(\mathbf{L})=(2+\beta)\mathbf{I} - \mathbf{L}_k^u, h(\lambda)=2+\beta - \lambda$, which is a adaptive low-pass hypergraph filter. Then for the low-pass user hypergraph filter, we have the following layer-wise propagation rule:

\begin{equation}
\begin{aligned}
\mathbf{U}_k^{(\ell+1)} &=\sigma\left(\mathbf{P}^{(l)}_k \mathbf{U}_k^{(\ell)} \mathbf{W}_k^{\mathrm{u},(\ell)} + \boldsymbol{b}\right)\\
\mathbf{P}^{(l)}_k &= \left(1 + \beta^{(l)}\right) \mathbf{I} + \mathbf{D}_u^{-\frac{1}{2}}\left(\mathbf{H}_k^{\mathrm{u}} \mathbf{P}_k^{\mathrm{v}} \mathbf{D}_{\mathrm{v}}^{-1} \mathbf{H}_k^{\mathrm{u}, T}-\widetilde{\mathbf{D}}_u\right) \mathbf{D}_u^{-\frac{1}{2}}, \\
\end{aligned}
\end{equation}

\subsection{Local-Global Bi-Directional Transfer Algorithm}

In this section, we introduce our proposed local-global bi-directional transfer algorithm. As shown in algorithm \ref{alg1}, in each round, the server sends the current global low-pass filter parameters $\boldsymbol{\phi}^{\mathrm{u}, (t)}$ and global domain-shared user representations $\mathbf{U}^{\mathrm{g}, (t)}$ to clients. In the local training stage, client $k$ updates its local high-pass hypergraph filter $\text{HHF}_k$ and low-pass hypergraph filter $\text{LHF}_k$ in an alternating manner, to perform knowledge transfer of global $\rightarrow$ local and local $\rightarrow$ global, as shown in lines $5-10$ of the pseudocode. This process allows for a trade-off between global and local models.

Finally, the server receives the updated local low-pass filter parameters $\{\widehat{\boldsymbol{\phi}}^{\mathrm{u}, (t+1)}_k\}^K_{k=1}$ and local domain-shared representations $\{\mathbf{U}^{\mathrm{s}, (t+1)}_k\}^K_{k=1}$ from clients, and updates the global model and user representations using weighted averaging.

\begin{algorithm2e}[!ht]
    \caption{Local-Global Bi-Directional Transfer Algorithm}
    \label{alg1}
    \KwIn{Local datasets $\mathcal{D}=\{\mathcal{D}_k\}^K_{k=1}$ (where $\mathcal{D}_k = (\mathcal{U}, \mathcal{V}_k, \mathcal{E}_k)$), local user/item hypergraph adjacency matrix $\mathbf{A}^{\text{u}}_k / \mathbf{A}^{\text{v}}_k$, total training round $T$}
    \KwOut{The optimal hypergraph high-pass filters $\{\boldsymbol{\theta}_k\}^K_{k=1}=\{(\boldsymbol{\theta}_k^{\mathrm{u}}, \boldsymbol{\theta}_k^{\mathrm{v}})\}^K_{k=1}$, \\ \quad The optimal hypergraph low-pass filters $\{\boldsymbol{\phi}_k\}^K_{k=1}=\{(\boldsymbol{\phi}_k^{\mathrm{u}}, \boldsymbol{\phi}_k^{\mathrm{v}})\}^K_{k=1}$
    }
     Server initialize $\boldsymbol{\phi}^{\mathrm{u}, (0)}$\;
     \For{round $t = 0, 1, \cdots T - 1$}{
        \For{each client $k \in K$ in parallel}{
        Receive $\boldsymbol{\phi}^{\mathrm{u}, (t)}_k$ and $\mathbf{U}^{\mathrm{g}, (t)}_k$ from server\;
        /* Global $\rightarrow$ Local */\\
        $\mathbf{U}_k^{\mathrm{e}}, \mathbf{V}_k^{\mathrm{e}}=\mathrm{HHF}_k\left(\mathbf{A}_k^{\mathrm{u}}, \mathbf{A}_k^{\mathrm{v}} ; \boldsymbol{\theta_k}\right)$\;
        $\left(\widehat{\boldsymbol{\theta}}_k^{\mathrm{u}}, \widehat{\boldsymbol{\theta}}_k^{\mathrm{v}}\right)=\arg \min _{\boldsymbol{\theta}_k} \mathcal{L}_k^{\mathrm{H}}\left(\mathbf{U}_k^{\mathrm{e}}, \mathbf{V}_k^{\mathrm{e}}\right) - \lambda I\left(\mathbf{U}_k^{\mathrm{e}}, \mathbf{U}_k^{\mathrm{g}, (t)}\right)$\;
        /* Local $\rightarrow$ Global */\\
        $\mathbf{U}_k^{\mathrm{s}}, \mathbf{V}_k^{\mathrm{s}}=\operatorname{LHF}_k\left(\mathbf{A}_k^{\mathrm{u}}, \mathbf{A}_k^{\mathrm{v}} ; \boldsymbol{\phi}_k\right)$\;
        $\left(\widehat{\boldsymbol{\phi}}_k^{\mathrm{u}}, \widehat{\boldsymbol{\phi}}_k^{\mathrm{v}}\right)=\arg \min _{\boldsymbol{\phi}_k} \mathcal{L}_k^{\mathrm{L}}\left(\mathbf{U}_k^{\mathrm{s}}, \mathbf{V}_k^{\mathrm{s}}\right) - \lambda I \left(\mathbf{U}_k^{\mathrm{s}}, \mathbf{U}_k^{\mathrm{e}}\right)$\;            
        Client send $\widehat{\boldsymbol{\phi}}^{\mathrm{u}}_k$ and $\mathbf{U}^{\mathrm{s}}_k$ to server\; 
        }
        $\boldsymbol{\phi}^{\mathrm{u}, (t+1)}=\sum_{k=1}^K \frac{\left|\mathcal{E}_{k}\right|}{\left|\mathcal{E}\right|} \widehat{\boldsymbol{\phi}}_k^{\mathrm{u}}$\;
        $\boldsymbol{\mathbf{U}}^{\mathrm{g}, (t+1)}=\sum_{k=1}^K \frac{\left|\mathcal{E}_{k}\right|}{\left|\mathcal{E}\right|} \mathbf{U}_k^{\mathrm{s}}$\;
    }
\end{algorithm2e}

\subsubsection{Loss for $\text{HHF}$ and $\text{LHF}$.} We adopt multi-class cross-entropy loss to train local high-pass
hypergraph filters and low-pass hypergraph filters. The loss of high-pass hypergraph filters is defined as follows (the same is true for low-pass hypergraph filters):

\begin{equation}
\mathcal{L}_k^{\mathrm{H}}=-\sum_{\left(u_i, v_j\right) \in \mathcal{E}_k} \log \frac{e^{s_k\left(\boldsymbol{u}_{k, i}, \boldsymbol{v}_{k, j}\right)}}{\sum_{v_j\in\mathcal{V}_k} e^{s_k\left(\boldsymbol{u}_{k, i}, \boldsymbol{v}_{k, j}\right)}}
\end{equation}

To reduce the computational complexity, we use the negative sampling method for training:

\begin{equation}
\mathcal{L}_k^{\mathrm{H}}=-\sum_{\left(u_i, v_j\right) \in \mathcal{E}_k}\left[\log \sigma\left(s_k\left(\boldsymbol{u}_{k, i}, \boldsymbol{v}_{k, j}\right)\right)+\mathbb{E}_{v_j^{\prime} \sim p^{\prime}_i(\mathcal{V}_k)} \log \sigma\left(-s_k\left(\boldsymbol{u}_{k, i}, \boldsymbol{v}_{k, j^{\prime}}\right)\right)\right],
\end{equation}

\noindent where $(u_i, v_j)/(u_i, v_j^{\prime})$ are positive/negative user-item interaction pairs, and $\boldsymbol{u}_{k, i}$, $\boldsymbol{v}_{k, j}$ and $\boldsymbol{v}_{k, j}^{\prime}$ are the corresponding user/item representations. Here $s_k(\cdot, \cdot)$ is the score function. $p^{\prime}_i(\mathcal{V}_k)$ denotes the nagative sampling distribution over the local items set $\mathcal{V}_k$ for the $i$-th user.

\subsubsection{MI for Knowledge Transfer.} For domain $k$, we adopt mutual information for knowledge transfer between local domain-exclusive and global domain-shared user representations during local training, which is computed as follows (the same applies to local domain-shared and local domain-exclusive user representations):

\begin{equation}
\begin{aligned}
&I\left(\mathbf{U}_k^{\mathrm{e}}, \mathbf{U}_k^{\mathrm{g}}\right) \\
&:= \sum_{u_i\in \mathcal{U}}\left[-\text{sp}\left( - \mathcal{D}_k \left(\boldsymbol{u}^{\text{e}}_{k, i}, \boldsymbol{u}^{\text{g}}_{k, i}\right)\right)\right] - \sum_{u_i, u_i^{\prime}\in\mathcal{U}, u_i^{\prime} \neq u_i}\left[\text{sp}\left(\mathcal{D}_k \left(\boldsymbol{u}^{\text{e}, \prime}_{k, i}, \boldsymbol{u}^{\text{g}}_{k, i}\right)\right)\right],
\end{aligned}
\end{equation}

\noindent where $\boldsymbol{u}^{\text{e}}_i$, $\boldsymbol{u}^{\text{g}}_i$ are the local domain exclusive representation and global domain shared representation of user $i$ respectively. $\mathcal{D}_k$ is the discriminator, $\text{sp}(x) = \log(1 + e^x)$ is the softplus function, and $\boldsymbol{u}^{\text{e}, \prime}_i$ denotes the negative sample randomly sampled from the user set $\mathcal{U}$.


\subsection{Hypergraph Contrastive Loss}

To better obtain domain-invariant user relationship information, when updating the low-pass user hypergraph filter, we do perturbation to the user hypergraph by randomly dropping a portion of edges. First, we sample a edge random masking matrix $\mathbf{B}_k^{\text{u}} \in \{0, 1\}^{|\mathcal{U}|\times|\mathcal{U}|}$, where $\left(\mathbf{B}_k\right)_{ij}\sim \text{Bernoulli}(1- p^{\text{d}})$ indicates whether to drop the edge between nodes $i$ and $j$. Here $p^{\text{d}}$ is the probability of each edge being dropped. Then the perturbed user hypergraph adjacency matrix can be computed as:

\begin{equation}
\begin{aligned}
\widetilde{\mathbf{A}}^{\text{u}}_k = \mathbf{A}^{\text{u}}_k \odot \mathbf{B}^{\text{u}}_k,
\end{aligned}
\end{equation}

\noindent where $\odot$ is the Hadamard product. Then, we compute the hypergraph contrastive loss as follows:

\begin{equation}
\begin{aligned}
\mathcal{L}_{k}^{\text{GCL}}\left(\mathbf{A}_k^{\mathrm{u}}, \widetilde{\mathbf{A}}_k^{\mathrm{u}}, \boldsymbol{\phi}^{\text{u}}_k\right)=-\sum_{u_i\in \mathcal{U}}\left(\ell^{\text{MI}}_k \left(\boldsymbol{u}_{k, i}^{\mathrm{s}}, \boldsymbol{z}_k^{\mathrm{s}}\right)+\ell^{\text{MI}}_k \left(\boldsymbol{z}_k^{\mathrm{s}}, \boldsymbol{u}_{k, i}^{\mathrm{s}}\right)\right),
\end{aligned}
\end{equation}

\noindent where graph representations $\boldsymbol{z}_k^s$ is aggregated through the readout function $g(\cdot)$ on the user hypergraph: 

\begin{equation}
\begin{aligned}
\quad \boldsymbol{z}_k^s=g\left(\{\boldsymbol{u}_{k, i}^{\text{s}}\}_{u_i\in \mathcal{U}}\right),
\end{aligned}
\end{equation}

\noindent and the contrastive infomax item $\ell^{\text{MI}}_k \left(\boldsymbol{u}_{k, i}^{\mathrm{s}}, \boldsymbol{z}_k^{\mathrm{s}}\right)$ is defined as follows:

\begin{equation}
\begin{aligned}
\ell^{\text{MI}}_k \left(\boldsymbol{u}_{k,i}^{\text{s}}, \boldsymbol{z}^{\text{s}}_k\right) = \mathbb{E}_{\mathbf{A}^{\text{u}}_k}\left[\log \mathcal{D}_k \left(\boldsymbol{u}^{\text{s}}_{k, i}, \boldsymbol{z}^{\text{s}}_k\right)\right] + \mathbb{E}_{\widetilde{\mathbf{A}}^{\text{u}}_k}\left[\log \left(1 - \mathcal{D}_k \left(\widetilde{\boldsymbol{u}}^{\text{s}}_{k, i}, \boldsymbol{z}^{\text{s}}_k\right)\right)\right],
\end{aligned}
\end{equation}

\noindent where $\mathcal{D}_k$ denotes the discriminator, which is trained to classify node embeddings based on whether they belong to the original graph $\mathbf{A}^{\text{u}}_k$ or the perturbed graph $\widetilde{\mathbf{A}}^{\text{u}}_k$. This loss can enforce the model to generate node embeddings that can distinguish between the real graph and its perturbed counterpart. Finally, the loss of the low-pass hypergraph filter in client $k$ can be denoted as


\begin{equation}
\begin{aligned}
\widetilde{\mathcal{L}}_k^{\mathrm{L}} = \mathcal{L}_k^{\mathrm{L}} + \gamma \mathcal{L}_{k}^{\text{GCL}}
\end{aligned}
\end{equation}

\section{Experiments}

In this section, we conduct a comprehensive set of experiments to evaluate the effectiveness of our framework FedHCDR by answering the following questions:

\begin{itemize}
\item \textbf{RQ1}: Does FedHCDR outperform state-of-the-art methods for FedCDR?
\item \textbf{RQ2}: How do HSD and HCL components enhance the performance of recommendations?
\item \textbf{RQ3}: Is our proposed HSD method able to achieve desirable decoupling?
\item \textbf{RQ4}: Does a global-local trade-off exist in the FedCDR scenario? How do different hyperparameters $\lambda$ and $\gamma$ impact the recommendation performance?
\end{itemize}

\begin{table}[!ht]
\renewcommand{\arraystretch}{1.2}
\centering
\small
\caption{\textbf{Statistics of Three FedCDR scenarios.}}
\label{table1}

\resizebox{0.7\linewidth}{!}{
\begin{tabular}{ccccccccc}

\hline
\textbf{Domain} & \textbf{\#Users} & \textbf{\#Items} & \textbf{\#Train} & \textbf{\#Valid} & \textbf{\#Test} & \textbf{Density} \\
\hline
Food            & 1898           & 11880  & 36097        & 1898           & 1898 & 0.177\%   \\
Kitchen        & 1898           &     18828                 &             44021                  &        1898 &
1898                          &       0.134\%          \\ 
Clothing          & 1898           & 16546  & 20919            & 1898           & 1898 & 0.079\% \\
Beauty    & 1898           &        12023                  &     30067        &      1898      &           1898            &   0.148\%
 \\ 
\hline
Sports  & 4004             & 35567 & 68627            & 4004           & 4004 & 0.054\% \\
Clothing      & 4004 &   24130                      & 31910            &                          4004 &       4004     &   0.041\%         &                       & \\
Elec     & 4004 &    50838               & 131107            &     4004                     &    4004        &   0.068\%         &                       & \\
Cell     & 4004 &    20556               & 36920            &     4004                     &    4004        &   0.055\%         &                       & \\
\hline
Sport     & 6657 &      46670                   & 108511            &                    6657      &    6657        &   0.039\%         &                       & \\
Garden     & 6657 &    24575                    & 69009            & 6657                         &    6657        &   0.050\%        &                       & \\
Home     & 6657 &   39426                      & 107273            &  6657                        &    6657        &   0.046\%      &                       & \\
Toys     & 6657 &   40406                      & 107041            &  6657                        &    6657        &   0.045\%      &                       & \\
\hline
\end{tabular}}
\end{table}

\subsection{Experimental Setup}

\subsubsection{Datasets.} We utilize publicly available datasets from the Amazon website $\footnote{\href{https://jmcauley.ucsd.edu/data/amazon/}{https://jmcauley.ucsd.edu/data/amazon/}}$ to construct FedCDR scenarios. Ten domains were selected to generate three cross-domain scenarios: Food-Kitchen-Cloth-Beauty (FKCB), Sports-Clothing-Elec-Cell (SCEC), and Sports-Garden-Home-Toys (SGHT). Following the approach of previous studies~\cite{BiTGCF, CDRIB}, we filter out users with less than 5 interactions and items with less than 10 interactions. For the dataset split, we follow the leave-one-out evaluation method employed in previous studies~\cite{BiTGCF, CDRIB}. Specifically, we randomly select two samples from each user's interaction history as the validation set and the test set, while the rest of the samples are used for training. The statistics of the FedCDR scenarios are summarized in Table~\ref{table1}.

\subsubsection{Evaluation Metrics.} To guarantee an unbiased evaluation, we follow the method described in Rendle's work~\cite{Sample}. Specifically, for each validation or test sample, we calculate its score along with 999 negative items. Subsequently, we evaluate the performance of the Top-K recommendation by analyzing the ranked list of 1,000 items using metrics such as MRR (Mean Reciprocal Rank)~\cite{MRR}, NDCG@10 (Normalized Discounted Cumulative Gain)~\cite{NDCG}, and HR@10 (Hit Ratio).

\subsubsection{Compared Baselines.} We compare our methods with two types of recommendation models: (1) single-domain recommendation methods, like NeuMF\cite{NeuMF}, LightGCN\cite{LightGCN}, DHCF\cite{DHCF}. (2) federated cross-domain recommendation methods, such as FedGNN\cite{FedGNN}, PriCDR\cite{PriCDR}, P2FCDR\cite{P2FCDR} and FPPDM++\cite{FPPDM++}.

\subsubsection{Implementation and Hyperparameter Setting.}

For all methods, the common hyperparameters are as follows: the training round is set to 60, the local epoch per client is set to 3, the early stopping patience is set to 5, the mini-batch size is set to 1024, the learning rate is set to 0.001, and the dropout rate is set to 0.3.

\subsection{Performance Comparisons (RQ1)}

Table~\ref{table2},~\ref{table3},~\ref{table4} present the performance of compared methods on three different FedCDR scenarios: Food-Kitchen-Clothing-Beauty, Sports-Clothing-Elec-Cell, and Sports-Garden-Home-Toys.

\begin{table*}[!ht]
\centering
\Large
\captionsetup{skip=0.1cm}
\caption{Federated experimental results(\%) on the FKCB scenario. Avg denotes the average result calculated from all domains. The best results are boldfaced.}
\label{table2}
\resizebox{\linewidth}{!}{%
\begin{tabular}{c|ccc|ccc|ccc|ccc|ccc} 
\hline
\multirow{3}{0.137\linewidth}{Method} & \multicolumn{3}{c|}{Food}       & \multicolumn{3}{c|}{Kitchen}  & \multicolumn{3}{c|}{Clothing}    & \multicolumn{3}{c|}{Beauty}    & \multicolumn{3}{c}{Avg}     \\ 
\cline{2-16}
                                                  & \multirow{2}{0.092\linewidth}{MRR} & HR   & NDCG & \multirow{2}{0.092\linewidth}{MRR} & HR   & NDCG & \multirow{2}{0.092\linewidth}{MRR} & HR    & NDCG  & \multirow{2}{0.092\linewidth}{MRR} & HR    & NDCG & \multirow{2}{0.092\linewidth}{MRR} & HR    & NDCG  \\ 
\cline{3-4}\cline{6-7}\cline{9-10}\cline{12-13}\cline{15-16}
                                                  &                                                & @10   & @10  &                                               & @10  & @10  &                                                & @10   & @10   &                                                & @10   & @10  &                                                & @10   & @10   \\ 
\hline
NeuMF                                 & 5.79                                          & 12.96  & 6.61 & 3.56                                          & 7.27 & 3.80
  & 1.61                                          & 2.32 & 1.43 & 4.22                                            & 9.11  & 4.64 & 3.80                                           & 7.92   & 4.12  \\ 
LightGCN                                 & 7.20                                          & 14.12  & 7.85 & 4.16                                          & 8.85 & 4.42
& 3.37                                          & 6.11 & 3.52 & 4.87                                            & 10.33  & 5.19 & 4.90                                           & 9.85   & 5.24  \\ 
DHCF                                 & 7.02                                          & 14.93  & 7.81 & 4.17                                          & 9.43 & 4.61
  & 3.58                                          & 6.59 & 3.79 & 4.98                                            & 10.57  & 5.43 & 4.93                                           & 10.38   & 5.41  \\ 
\hline
FedGNN                              & 7.15                                           & 13.91 & 7.75 & 4.15                                         & 9.01 & 4.46 & 3.45                                          & 6.38 & 3.65 & 4.86                                           & 10.12 & 5.12 & 4.90                                           & 9.85  & 5.25 \\
PriCDR                              & 7.34                                           & 16.60 & 8.58 & 4.55                                         & 9.11 & 4.88 & 3.49                                          & 5.95 & 3.55 & 5.26                                           & 10.48 & 5.51 & 5.16                                           & 10.54  & 5.63  \\ 
P2FCDR                                 & 7.08                                           & 13.91 & 7.68 & 4.28                                          & 8.96 & 4.63 & 3.18                                           & 6.53  & 3.51 & 4.27                                            & 9.64  & 4.53 & 4.70                                           & 9.76  & 5.09  \\ 
FPPDM++                                  & 7.25                                          & 14.01 & 7.85 & 4.19                                          & 9.17 & 4.54 & 3.60                                          & 6.27  & 3.71 & 4.89                                           & 10.22  & 5.16 & 4.98                                           & 9.92  & 5.31  \\

\hline
\textbf{FedHCDR(Ours)}                                     & \textbf{7.35}                                           & \textbf{16.75} & \textbf{8.62}   & \textbf{4.56}                                         & \textbf{9.69} & \textbf{4.97} & \textbf{3.72}                                          & \textbf{6.61} & \textbf{4.01} & \textbf{5.32}                                             & \textbf{10.59}  & \textbf{5.56} & \textbf{5.24}                                             & \textbf{11.00} & \textbf{5.79}  \\
\hline
\end{tabular}
}
\end{table*}

\begin{table*}[!ht]
\centering
\Large
\captionsetup{skip=0.1cm}
\caption{Federated experimental results(\%) on the SCEC scenario. Avg denotes the average result calculated from all domains. The best results are boldfaced.}
\label{table3}
\resizebox{\linewidth}{!}{%
\begin{tabular}{c|ccc|ccc|ccc|ccc|ccc} 
\hline
\multirow{3}{0.137\linewidth}{Method} & \multicolumn{3}{c|}{Sports}       & \multicolumn{3}{c|}{Clothing}  & \multicolumn{3}{c|}{Elec}    & \multicolumn{3}{c|}{Cell}    & \multicolumn{3}{c}{Avg}     \\ 
\cline{2-16}
                                                  & \multirow{2}{0.092\linewidth}{MRR} & HR   & NDCG & \multirow{2}{0.092\linewidth}{MRR} & HR   & NDCG & \multirow{2}{0.092\linewidth}{MRR} & HR    & NDCG  & \multirow{2}{0.092\linewidth}{MRR} & HR    & NDCG & \multirow{2}{0.092\linewidth}{MRR} & HR    & NDCG  \\ 
\cline{3-4}\cline{6-7}\cline{9-10}\cline{12-13}\cline{15-16}
                                                  &                                                & @10   & @10  &                                               & @10  & @10  &                                                & @10   & @10   &                                                & @10   & @10  &                                                & @10   & @10   \\ 
\hline
NeuMF                                 & 2.30                                          & 3.90  & 2.19 & 1.06                                          & 1.70 & 0.86
  & 4.68                                          & 9.42 & 5.13 & 3.14                                            & 5.57  & 3.21 & 2.80                                           & 5.14   & 2.85  \\
LightGCN                                 & 4.03                                          & 8.64   & 4.43 & 3.37                                          & 6.82 & 3.61
& 6.78                                          & 12.86 & 7.43 & 5.09                                            & 10.21  & 5.61 & 4.82                                           & 9.63   & 5.27  \\
DHCF                                 & 3.92                                          & 8.12  & 4.22 & 3.36                                          & 6.89 & 3.62
  & 6.68                                          & 12.96 & 7.39 & 4.98                                            & 10.31  & 5.57 & 4.74                                           & 9.57   & 5.20  \\ 
\hline
FedGNN                              & 3.99                                           & 8.57 & 4.38 & 3.38                                         & 6.87 & 3.63 & 6.78                                          & 13.04 & 7.47 & 5.12                                           & 10.24 & 5.64 & 4.82                                           & 9.68  & 5.28  \\ 
PriCDR                                 & 3.89                                           & 8.04  & 4.19 & 3.40                                         & 6.82 & 3.66 & 5.50                                           & 10.76  & 6.05 & 5.34                                            & 10.74  & 5.93 & 4.53                                           & 9.09  & 4.96  \\ 
P2FCDR                                  & 3.77                                          & 7.84 & 4.06 & 3.13                                          & 6.59 & 3.41 & 6.25                                          & 12.36  & 6.93 & 5.14                                           & 10.51  & 5.75 & 4.57                                           & 9.33  & 5.04  \\
FPPDM++                                  & 4.06                                          & 8.64 & 4.45 & 3.38                                          & 6.74 & 3.60 & 6.71                                          & 12.99  & 7.39 & 5.16                                           & 10.41  & 5.72 & 4.83                                           & 9.70  & 5.29  \\
\hline
\textbf{FedHCDR(Ours)}                                     & \textbf{4.47}                                           & \textbf{9.04} & \textbf{4.90}   & \textbf{3.45}                                         & \textbf{6.92} & \textbf{3.66} & \textbf{6.88}                                          & \textbf{13.61} & \textbf{7.65} & \textbf{5.75}                                             & \textbf{11.11}  & \textbf{6.31} & \textbf{5.14}                                             & \textbf{10.17} & \textbf{5.63}  \\
\hline
\end{tabular}
}
\end{table*}

\begin{table*}[!ht]
\centering
\Large
\captionsetup{skip=0.1cm}
\caption{Federated experimental results(\%) on the SGHT scenario. Avg denotes the average result calculated from all domains. The best results are boldfaced.}
\label{table4}
\resizebox{\linewidth}{!}{%
\begin{tabular}{c|ccc|ccc|ccc|ccc|ccc} 
\hline
\multirow{3}{0.137\linewidth}{Method} & \multicolumn{3}{c|}{Sports}       & \multicolumn{3}{c|}{Garden}  & \multicolumn{3}{c|}{Home}    & \multicolumn{3}{c|}{Toys}    & \multicolumn{3}{c}{Avg}     \\ 
\cline{2-16}
                                                  & \multirow{2}{0.092\linewidth}{MRR} & HR   & NDCG & \multirow{2}{0.092\linewidth}{MRR} & HR   & NDCG & \multirow{2}{0.092\linewidth}{MRR} & HR    & NDCG  & \multirow{2}{0.092\linewidth}{MRR} & HR    & NDCG & \multirow{2}{0.092\linewidth}{MRR} & HR    & NDCG  \\ 
\cline{3-4}\cline{6-7}\cline{9-10}\cline{12-13}\cline{15-16}
                                                  &                                                & @10   & @10  &                                               & @10  & @10  &                                                & @10   & @10   &                                                & @10   & @10  &                                                & @10   & @10   \\ 
\hline
NeuMF                                 & 3.27                                          & 6.46  & 3.39 & 4.39                                          & 8.49 & 4.59
  & 4.23                                          & 8.35 & 4.50 & 2.54                                            & 4.81  & 2.45 & 3.61                                           & 7.03   & 3.73  \\ 
LightGCN                                 & 4.61                                          & 8.74  & 4.76 & 5.45                                          & 10.33 & 5.67
& 5.83                                          & 11.03 & 6.20 & 2.95                                            & 5.14  & 2.78 & 4.71                                           & 8.81   & 4.85  \\
DHCF                                 & 4.78                                          & 8.92  & 4.95 & 5.49                                          & 10.43 & 5.73
& 5.82                                          & 11.31 & 6.28 & 3.34                                            & 5.35  & 3.14 & 4.86                                           & 9.00   & 5.03  \\ 
\hline
FedGNN                              & 4.72                                           & 8.74 & 4.84 & 5.60                                          & 10.61 & 5.86 & 5.91                                          & 11.30 & 6.33 & 3.15                                           & 5.06 & 2.91 & 4.84                                           & 8.93  & 4.99  \\
PriCDR                              & 4.59                                           & 8.46 & 4.78 & 6.02                                         & 11.69 & 6.46 & 5.56                                          & 11.09 & 6.20 & 4.92                                           & \textbf{9.15} & 5.02 & 5.27                                           & 10.09  & 5.61  \\
P2FCDR                                 & 5.06                                           & 9.22 & 5.25 & 5.83                                          & 11.06 & 6.16 & 5.94                                           & 11.31  & 6.36 & 4.14                                            & 6.41  & 4.02 & 5.24                                           & 9.50  & 5.45  \\
FPPDM++                                  & 4.60                                          & 8.73 & 4.75 & 5.49                                          & 10.32 & 5.70 & 5.80                                          & 11.15  & 6.20 & 3.12                                           & 5.20  & 2.92 & 4.75                                           & 8.85  & 4.89  \\
\hline
\textbf{FedHCDR(Ours)}                                     & \textbf{5.46}                                           & \textbf{10.08} & \textbf{5.75}   & \textbf{6.13}                                         & \textbf{11.85} & \textbf{6.58} & \textbf{6.18}                                          & \textbf{12.15} & \textbf{6.74} & \textbf{4.92}                                             & 8.28  & \textbf{5.04} & \textbf{5.67}                                             & \textbf{10.59} & \textbf{6.03}  \\
\hline
\end{tabular}
}
\end{table*}

Based on the experimental results, several insightful observations can be made: (1) Among the single domain baselines, LightGCN and DHCF perform better than NeuMF. This finding validates that modeling the relationship between users and items by GCN can enhance the representations in the FedCDR scenario. (2) Most cross-domain baselines perform better than single-domain baselines, which indicates that cross-domain knowledge helps improve recommendation performance. (3) Among the cross-domain baselines, both PriCDR and FPPDM++ outperform FedGNN and P2FCDR in most cases, indicating that representation/distribution alignment can effectively accomplish the knowledge transfer between domains in the FedCDR scenario. (4) Our proposed method, FedHCDR, significantly outperforms all baselines in multiple metrics. This emphasizes the crucial role of hypergraph signal decoupling and hypergraph contrastive learning in capturing both local and global user features.

\subsection{Ablation Study (RQ2)}

We conduct an ablation study on the performance of FedHCDR, specifically examining the impact of HSD and HCL. Table~\ref{table5} presents the performance results of different model variants in three FedCDR scenarios. LocalHF represents the HF model (vanilla hypergraph filter) without federated aggregation, FedHCDR-w/o (HSD, HCL) corresponds to FedHCDR without HSD and HCL, and FedHCDR-w/o HCL refers to FedHCDR without HCL. It is evident from the findings that FedHCDR-w/o (HSD, HCL) occasionally performs worse than LocalHF, highlighting the significance of data heterogeneity. Interestingly, FedHCDR-w/o HCL greatly outperforms both LocalHF and FedHCDR-w/o (HSD, HCL), indicating the effectiveness of HSD in addressing the data heterogeneity across domains. Furthermore, the utilization of HCL enables further improvements in model performance.

\begin{table}
\centering
\Large
\caption{Ablation study on FKCB, SCEC, and SGHT scenarios.}
\label{table5}
\resizebox{0.8\linewidth}{!}{%
\begin{tabular}{c|ccc|ccc|ccc} 
\hline
\multirow{3}{0.142\linewidth}{Method} & \multicolumn{3}{c|}{FKCB} & \multicolumn{3}{c|}{SCEC} & \multicolumn{3}{c}{SGHT}  \\ 
\cline{2-10}
                                                  & \multirow{2}{0.092\linewidth}{MRR} & HR & NDCG    & \multirow{2}{0.092\linewidth}{MRR} & HR & NDCG      & \multirow{2}{0.092\linewidth}{MRR} & HR & NDCG         \\ 
\cline{3-4}\cline{6-7}\cline{9-10}
                                                  &                                                & @10 & @10     &                                                & @10 & @10      &                                                & @10  & @ 10        \\ 
\hline
LocalHF                                   & 4.97                                           & 9.98  & 5.33 & 4.84                                           & 9.70  &  5.31  & 4.86                                          & 8.85  &  4.97   \\
FedHCDR - w/o (HSD, HCL)                            & 4.98                                          & 10.02 &  5.34 & 4.81                                          & 9.69  & 5.28   & 4.95                                         & 9.03 &  5.09     \\
FedHCDR - w/o HCL                                    & 5.18                                        & 10.95  &  5.73  & 5.07                                          & 10.12  & 5.60   & 5.64                                          & 10.58 & 6.01       \\ \textbf{FedHCDR(Ours)}
                                                  &        \textbf{5.24}                                        & \textbf{11.00}  &  \textbf{5.79}    & \textbf{5.14}                                              &      \textbf{10.17} &  \textbf{5.63}   &    \textbf{5.67}                                            & \textbf{10.59}    &  \textbf{6.03}       \\ 
\hline
\end{tabular}
}
\end{table}

\subsection{Discussion of the user representation (RQ3)}

\begin{figure*}[ht]
    \centering
    \subfigure[Rep. in FKCB scenario]{
    \includegraphics[width=2.1in, scale=0.7]{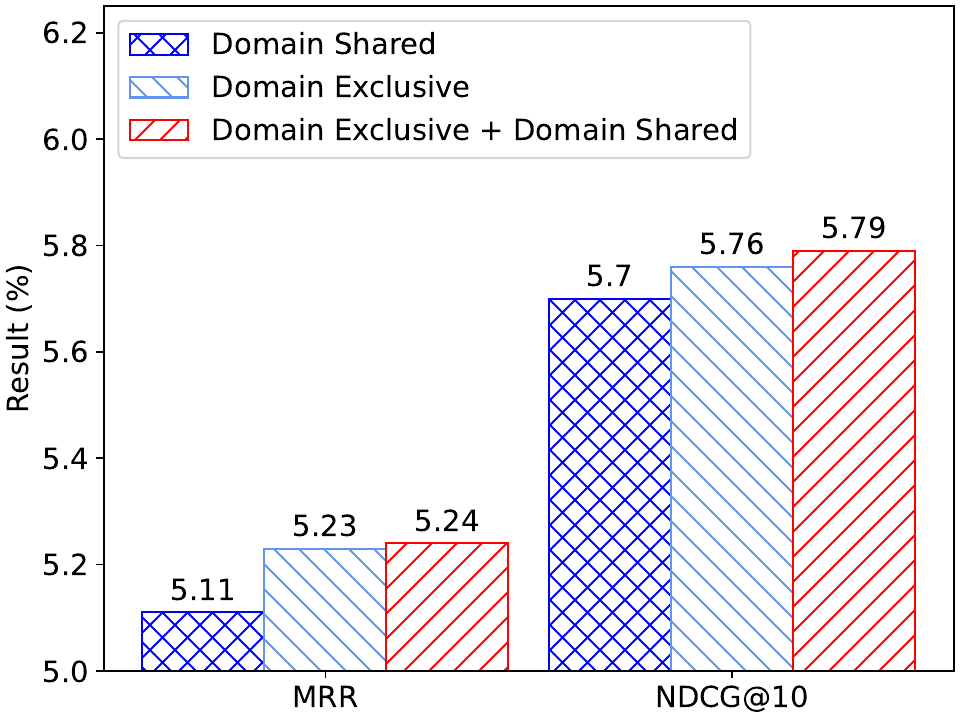}}
    \subfigure[Rep. in SGHT scenario]{
        \includegraphics[width=2.1in, scale=0.7]{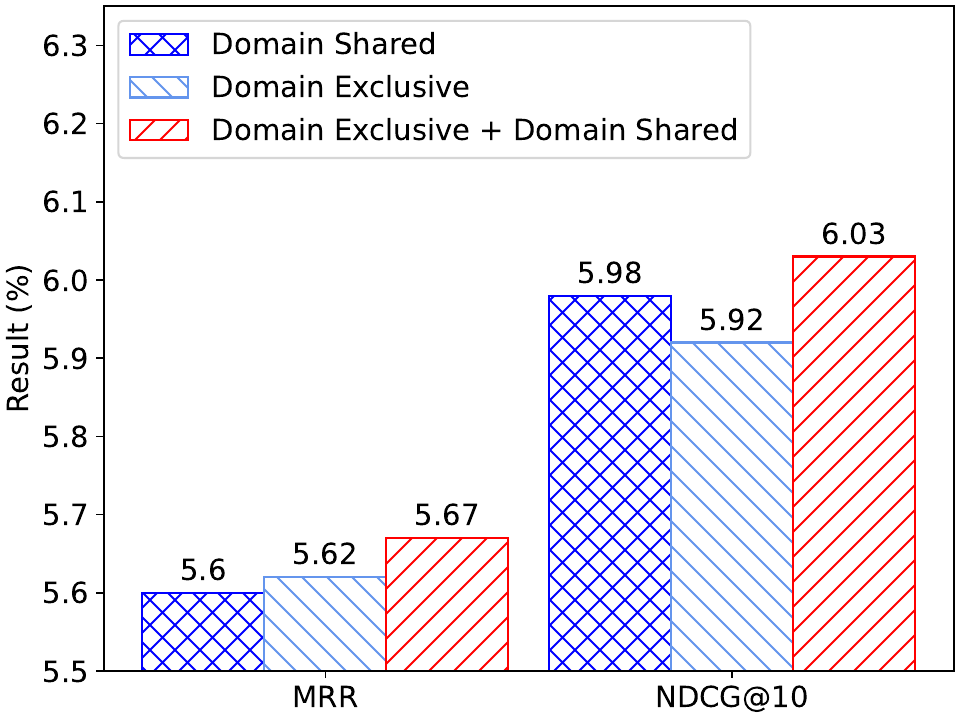}}
    \caption{The predictive results of representations in FKCB and SGHT scenario.}
    \label{fig5}
\end{figure*}

In this section, we aim to further validate the ability of our HSD to learn both domain-shared and domain-exclusive representations for users. To achieve this, we conduct a comparative analysis of three types of representations: domain-shared, domain-exclusive, and domain-exclusive + domain-shared representations. The predictive performance of these representations is compared, as illustrated in Fig.~\ref{fig5}. The results of our analysis reveal several interesting observations: (1) The predictive performance varies among the three types of representations, highlighting the effectiveness of our HSD. (2) The domain-exclusive + domain-shared representations outperform both the domain-shared and domain-exclusive representations, indicating that integrating information from multiple domains by considering both domain-shared and domain-exclusive features is highly effective.

\subsection{Influence of hyperparameters (RQ4)} 
\begin{figure*}[!ht]
    \centering
    \subfigure[Impact of coefficient $\lambda$]{
    \includegraphics[width=2.1in, scale=0.7]{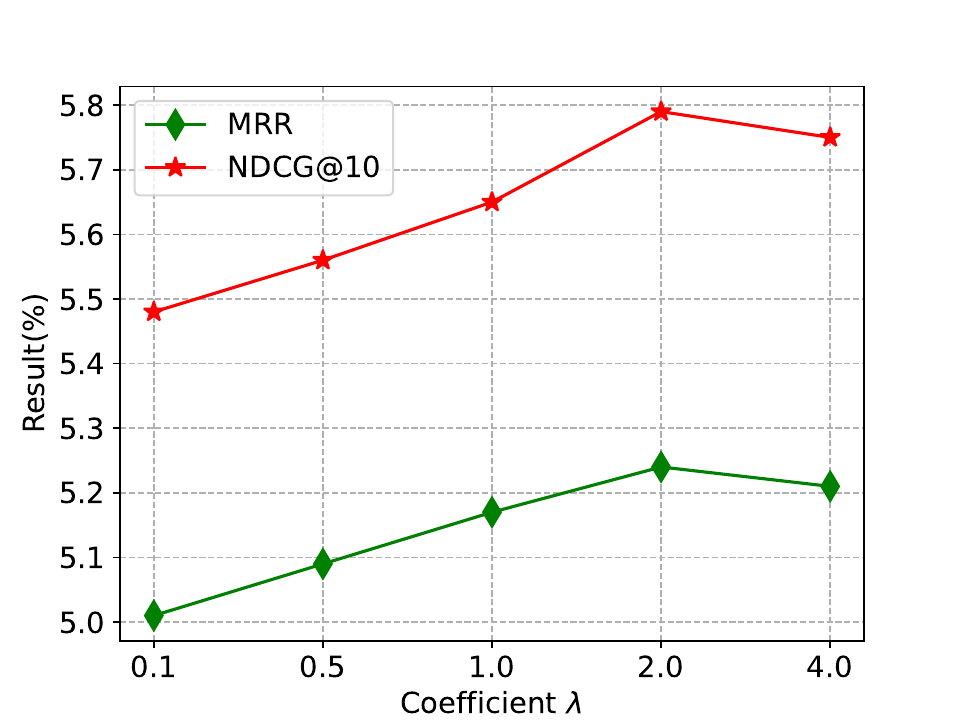}}
    \subfigure[Impact of coefficient $\gamma$]{
    \includegraphics[width=2.1in, scale=0.7]{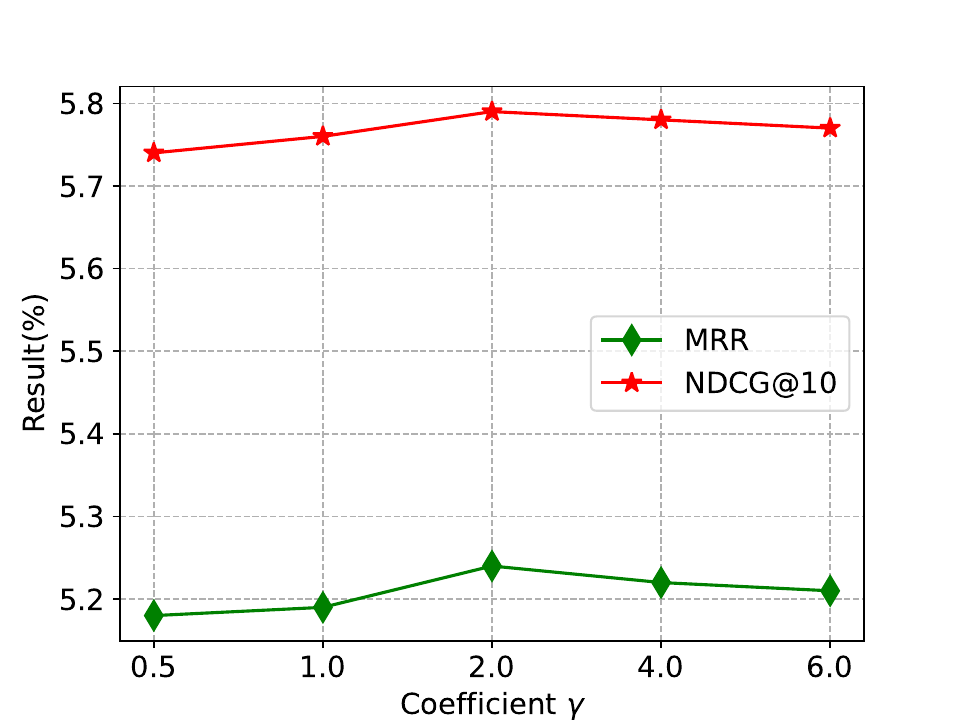}}
    \caption{Impact of coefficient $\lambda$ and $\gamma$.}
    \label{fig6}
\end{figure*}

Fig.~\ref{fig6} displays the performance of MRR and @NDCG@10 as the coefficients $\lambda$ and $\gamma$ increase. The following observations can be made: (1) The overall performance of FedHCDR initially increases and then decreases as $\lambda$ increases, peaking at 2.0. This suggests that an $\lambda$ coefficient of 2.0 is optimal for local-global bi-direction transfer and highlights the local-global trade-off. (2) The overall performance of FedHCDR follows a similar pattern with the increase of $\gamma$, reaching its peak at 2.0. This indicates that a $\gamma$ coefficient of 2.0 is optimal for hypergraph contrastive learning.

\section{Related Work}

\subsection{GCN-based Recommendation}

The development of graph neural networks has attracted considerable attention in the exploration of GCN-based Recommendation~\cite{GC_MC, SpectralCF}. NGCF~\cite{NGCF} leverages the user-item graph structure by propagating embeddings throughout it. LightGCN~\cite{LightGCN} simplifies the model design by including only the neighborhood aggregation for collaborative filtering. DHCF~\cite{DHCF} utilizes the hypergraph structure to model users and items, effectively capturing explicit hybrid high-order correlations. SGL and MCCLK~\cite{SGL, MCCLK} integrate contrastive learning into GCN-based recommendation methods. However, the aforementioned methods solely concentrate on a single domain, thus unable to fully exploit user data from multiple domains.

\subsection{Cross-Domain Recommendation}

DTCDR and DDTCDR~\cite{DTCDR, DDTCDR} enhance the performance of recommendations on dual-target domains simultaneously. BiTGCF~\cite{BiTGCF} introduces an innovative bi-directional transfer learning approach for cross-domain recommendation, utilizing the graph collaborative filtering network as the foundational model. DisAlig~\cite{DisAlign} proposes the use of Stein path alignment to align the latent embedding distributions across domains. CDRIB~\cite{CDRIB} suggests the use of information bottleneck regularizers to establish user-item correlations across domains. Nonetheless, these methods require access to all user-item interactions across domains, rendering them infeasible in the federated learning setting.

\subsection{Federated Cross-Domain Recommendation}
FedMF \cite{FedMF} effectively incorporates federated learning into the field of cross-domain recommendation. FedCTR \cite{FedCTR} proposes a framework for training a privacy-preserving CTR prediction model across multiple platforms.
FedCDR \cite{FedCDR} deploys the user personalization model on the client side and uploads other models to the server during aggregation. P2FCDR \cite{P2FCDR} proposes a privacy-preserving federated framework for dual-target cross-domain recommendation. FPPDM++ \cite{FPPDM++} presents a framework that models and shares the distribution of user/item preferences across various domains. Nevertheless, none of these methods address the issue of cross-domain data heterogeneity.

\section{Conclusion}
In this paper, we present a novel framework called FedHCDR, designed to enable domains to collaboratively train better performing CDR models while ensuring privacy protection. To address the issue of data heterogeneity, we introduce a hypergraph signal decoupling method called HSD that decouples user features into domain-exclusive and domain-shared features. Additionally, we devise a hypergraph contrastive learning module called HCL to learn more extensive domain-shared user relationship information by applying graph perturbation to the user hypergraph.

\begin{credits}
\subsubsection{\ackname} This work is supported by Guangdong Major Project of Basic and Applied Basic Researche (No. 2019B03030200245), the National Natural Science Foundation of China (62227808), and Shenzhen Science and Technology Program (Grant No.ZDSYS20210623091809029).
\end{credits}

%
%
%
%

\end{sloppypar}
\end{document}